\documentclass[10pt, a4paper]{article}
\usepackage{amsmath}
\usepackage{multirow}
\usepackage{url}
\usepackage{tikz}
\usetikzlibrary{shapes, arrows.meta, positioning}
\usepackage[ruled,vlined]{algorithm2e}
\usepackage{subcaption}
\usepackage{tabularx}
\usepackage[final]{lrec2026} 

\title{Efficient Topic Extraction via Graph-Based Labeling: A Lightweight Alternative to Deep Models}

\name{
\large \bfseries
Salma MEKAOUI\textsuperscript{1,*}, Hiba SOFYAN\textsuperscript{2}, Imane AMAAZ\textsuperscript{2}, Imane BENCHRIF\textsuperscript{2},\\
\large \bfseries Arsalane ZARGHILI\textsuperscript{3}, Ilham CHAKER\textsuperscript{3}, Nikola S. Nikolov\textsuperscript{1}\\
}

\address{\textsuperscript{1}University of Limerick, Department of Computer Science and Information Systems, Ireland\\
         \textsuperscript{2}Euromed University Of Fez, School of Digital Engineering and Artificial Intelligence, Morocco\\
		 \textsuperscript{3}Faculty of Sciences and Technology, University Sidi Mohamed Ben Abdellah, Morocco\\[1ex]
         \textsuperscript{1,*}Corresponding author: mekaoui.salma@ul.ie
         }

\abstract{
Extracting topics from text has become an essential task, especially with the rapid growth of unstructured textual data. Most existing works rely on highly computational methods to address this challenge. In this paper, we argue that probabilistic and statistical approaches, such as topic modeling (TM), can offer effective alternatives that require fewer computational resources. TM is a statistical method that automatically discovers topics in large collections of unlabeled text; however, it produces topics as distributions of representative words, which often lack clear interpretability. Our objective is to perform topic labeling by assigning meaningful labels to these sets of words. To achieve this without relying on computationally expensive models, we propose a graph-based approach that not only enriches topic words with semantically related terms but also explores the relationships among them. By analyzing these connections within the graph, we derive suitable labels that accurately capture each topic’s meaning. We present a comparative study between our proposed method and several benchmarks, including ChatGPT-3.5~\cite{openai2023chatgpt}, across two different datasets. Our method achieved consistently better results than traditional benchmarks in terms of BERTScore and cosine similarity and produced results comparable to ChatGPT-3.5, while remaining computationally efficient. Finally, we discuss future directions for topic labeling and highlight potential research avenues for enhancing interpretability and automation.
 \\ \newline \Keywords{Topic Modeling, Topic Labeling, Graph-based Methods, ConceptNet, Natural Language Processing} }

\begin{document}

\maketitleabstract

\section{Introduction}
Extracting topics from text is a rapidly growing area of research, driven by the enormous and continuously increasing volume of textual data across various domains. This surge has created a demand for methods that can accurately and efficiently identify the underlying themes within large corpora. Many recent studies focus on the application and advancement of Natural Language Processing (NLP) techniques, particularly leveraging sophisticated methods such as topic extraction, semantic understanding, and text organization, often using Large Language Models (LLMs)~\cite{wang2024improving,nakanishi2025fusion,bitaraf2024unveiling,reynolds2025research,liautomatic}. However, these approaches are typically computationally expensive and require substantial resources to achieve high-quality results.

In contrast, topic modeling (TM) provides a well-established and efficient alternative. TM is an unsupervised machine learning (ML) technique that automatically discovers latent topics in large collections of unlabelled text, without requiring prior training data~\cite{churchill2022evolution}. By identifying the underlying thematic structure of a corpus, TM enables researchers to generate representative topics in a resource-efficient manner. While TM can reveal the underlying structure of a corpus, a major limitation lies in its inability to directly produce human-readable topics~\cite{churchill2022evolution}. Typically, TM outputs  are a distribution of words associated with each topic, which may indicate the most relevant terms but often lacks interpretability for human readers. To address this challenge, topic labeling (TL) has been introduced. TL~\cite{mekaoui2025systematic} is an approach for assigning comprehensive and semantically meaningful labels to TM results, ensuring that they are interpretable and understandable from a human perspective. 
TL has gained increasing importance and has been applied across diverse domains. A notable application is presented by Nolasco~\emph{et al.}~\cite{nolasco2016detecting} who applied TL on a scholarly dataset derived from KDD conference proceedings (Knowledge Discovery and Data Mining), demonstrating that their method can effectively generate topic labels with efficiency comparable to human-generated labels. In~\cite{tiwari2023advanced}, TL was applied to extract topics from short texts. Similarly, Hagerer~\emph{et al.}~\cite{hagerer2021socialvistum} utilized TL on a dataset of online user comments about organic food, collected from multiple platforms such as Reddit, Quora, Disqus, and major news websites including The Washington Post and The New York Times.

Although topic labeling has been widely adopted for labeling topics in various domains, numerous studies emphasize persistent challenges and limitations in this field. A central concern relates to computational complexity. Neural-based approaches~\cite{ramon2023automatic,mao2016novel,tiwari2023advanced,mu2024large} show that resource-demanding models, such as ChatGPT, require considerable computational power and often suffer from limited accessibility. Several works point to methodological limitations in TL, including the inability to create novel labels, difficulties in selecting an appropriate model, and the tendency to produce labels that are either too generic or insufficiently discriminative for meaningful content interpretation~\cite{alokaili2020automatic,flondor2023analysing,ramon2023automatic,tang2016topic,nolasco2016detecting,sanjaya2018harnessing,kozbagarov2021new,li2016effective,li2019filtering}. Ontology and graph-related challenges are also widely recognized, as these approaches typically involve elaborate, multi-stage pipelines that integrate TM outputs into ontological or graph structures and align them with domain-specific concepts to derive suitable labels for TM results~\cite{allahyari2015automatic,hagerer2021socialvistum,adhitama2017topic,scelsi2021principled,chaudhary2024top2label}. Furthermore, studies involving human labeling highlight the substantial manual effort required in the labeling process~\cite{flondor2023analysing,sanjaya2018harnessing,zheng2022automatic}. Finally, issues of coherence, granularity, precision, and objectivity in label interpretation are consistently reported in the literature~\cite{mao2016novel,sanjaya2018harnessing,davoudi2015ontology,he2019automatic,gourru2018united,mao2012automatic}.

In this paper, we introduce a novel TL method that leverages graph structures to generate representative labels for sets of words which represent a topic. Our method addresses several limitations identified in previous studies. Specifically, we focus on integrating TM outputs into a graph structure and identifying the most representative nodes within that graph through a simpler and more transparent procedure. We further hypothesize that adopting a graph-based approach, rather than relying on computationally intensive neural models, can alleviate issues related to high resource demands, model selection complexity, and the tendency of existing methods to produce labels that are overly generic or insufficiently discriminative.

This paper is structured as follows. Section~\ref{sec:related_works} reviews the existing works on TL. Section~\ref{sec:Methodology} presents the methodology adopted in this study. Section~\ref{sec:Experiments} discusses and analyzes the experimental results, including comparisons with benchmark methods. Finally, Section~\ref{sec:conclusion} concludes the paper and outlines future research directions.
\section{Related Works}
\label{sec:related_works}
This section reviews the main approaches proposed in the literature for TL. Existing methods can be broadly categorized into four groups: traditional methods, neural-based methods, ontology and graph-based methods, and hybrid approaches.
\subsection{Traditional Methods}
Earlier attempts to solve the TL task mainly relied on classical machine learning (ML) algorithms and information retrieval (IR) strategies. These include well-known classifiers such as support vector machines (SVM)~\cite{effendi2023hybrid}, k-nearest neighbors (KNN)~\cite{mao2016novel}, in addition to extensions of TM methods like non-negative matrix factorization (NMF) and latent direchlet allocation (LDA)~\cite{li2022guided,zha2019multi,mao2015ehllda,liu2019label,wood2017source}. From the IR perspective, methods have been designed to extract candidate terms or phrases using a predefined dictionary of meaningful words~\cite{tang2016topic}, candidate selection algorithms~\cite{nolasco2016detecting}, or truth discovery frameworks~\cite{sanjaya2018harnessing}. Finally, some studies rely on human expertise for topic annotation. In these cases, domain experts or annotators validate and assign labels manually to the TM results~\cite{flondor2023analysing,korenvcic2015getting}. Traditional TL approaches remain limited by manual intervention, slow processing, and reliance on external knowledge sources, leading to inconsistent and less adaptive labels across datasets.
\subsection{Neural-based Methods}
With the rise of deep learning, neural models have gained increasing attention in TL research. Some studies directly employ large language models (LLMs) to generate suitable labels for TM results~\cite{ramon2023automatic,mu2024large}, while others adapt or fine-tune them for this specific task. For instance, the Bidirectional and Auto-Regressive Transformer (BART)~\cite{lewis2019bart} has been extended into a TL variant (BART-TL) through a weakly supervised fine-tuning process~\cite{popa2021bart}. Similarly, the Seq2Seq architecture has been applied using the Wikipedia dataset to produce meaningful topic labels~\cite{alokaili2020automatic}. Contextual embeddings have also been investigated, such as in the contextualized topic model (CTM)~\cite{zheng2022automatic}. Other approaches rely on word embedding strategies to enrich the semantics of the label~\cite{kozbagarov2021new,tiwari_advanced_2023}. Along the same line, neural embedding topic labeling (NETL) exploits neural embeddings, specifically word2vec~\cite{church2017word2vec} for word representations and doc2vec~\cite{mikolov2013distributed} for document/title embeddings, to identify the most relevant labels for each topic~\cite{bhatia2016automatic}. Neural-based approaches leverage deep architectures to capture semantic relationships beyond TM resulted words. However, they often require large annotated datasets, high computational resources, and struggle with multilingual adaptation and long-text processing.
\subsection{Ontology and Graph Based Methods}
Another area of research leverages ontologies to map TM results (typically word lists) to domain concepts and categories. This knowledge-driven approach has been applied in several studies, demonstrating its effectiveness in producing semantically meaningful topic labels~\cite{kinariwala2021onto_tml, allahyari2015automatic, adhitama2017topic, kim2019ontology, davoudi2015ontology, magatti2009automatic, slabbekoorn2016ontology, zosa2022multilingual}.

Building on this semantic perspective, graph-based representations have also been explored to capture structural relationships among topics or terms. Such approaches include social graph modeling~\cite{hagerer2021socialvistum}, graph-based ranking~\cite{he2019automatic}, semantic enrichment using the ConceptNet knowledge graph~\cite{chaudhary2024top2label}, and hierarchical topic construction with Bayesian Rose Trees (BRTs), which discover topic labels by organizing words and concepts into semantic hierarchies~\cite{ogawa2021text}. Ontology and graph-based methods leverage structured knowledge sources to enhance the interpretability of TL. However, they face challenges related to ontology construction, concept selection, and the accurate mapping of topic words to semantic hierarchies, which can be complex.
\subsection{Hybrid approach}
These methods begin by extracting relevant knowledge from external resources or specialized domain dictionaries, such as Wikipedia~\cite{vathi2017mining,hossain2019polynomial,scelsi2021principled,lau2011automatic}. This foundational knowledge is then combined with machine learning algorithms, neural networks, or other computational techniques to improve topic identification. For example, Gourru~\emph{et al.}~\cite{gourru2018united} demonstrate a hybrid approach by merging multiple unsupervised n-gram-based labelers. Similarly, Li~\emph{et al.}~\cite{li2016effective,li2019filtering} showcase the effectiveness of integrating various methods within their work. However, these hybrid approaches are often limited by evaluations on small or narrow datasets, raising concerns about their generalizability. Additionally, while combining multiple methods can improve coverage, issues such as label precision and the selection of the most appropriate labeler remain largely unresolved.
\section{Methodology}
\label{sec:Methodology}
\begin{figure}[!ht]
\centering
\tikzset{
  box/.style={
    rectangle, rounded corners, draw=black, align=left,
    text width=7.5cm, minimum height=0.4cm, minimum height=0.4cm, inner sep=3pt
  },
  arrow/.style={-{Latex[length=3mm]}, thick}
}

\begin{tikzpicture}[node distance=0.3cm]
\node[box, fill=blue!10] (s1) {\textbf{1. Input: Topic Modeling Results}\\
Input of word lists for each topic, e.g., [``obama'', ``mccain'', ``campaign'', ``john'', ``barack'', ``president'', ``senator'', ``candidate'', ``convention'', ``clinton''].};

\node[box, fill=green!10, below=of s1] (s2) {\textbf{2. Sentence Construction}\\
Transformation of word lists into sentences, e.g., ``obama, mccain, campaign, john, barack, president, senator, candidate, convention, clinton''.};

\node[box, fill=yellow!10, below=of s2] (s3) {\textbf{3. Sentence Embedding}\\
Generate embeddings of the constructed sentence using a pre-trained model.};

\node[box, fill=cyan!10, below=of s3] (s4) {\textbf{4. Label Selection Algorithm}\\
Apply the algorithm to select the best label based on similarity scores.};

\node[box, fill=purple!10, below=of s4] (s5) {\textbf{5. Output: Topic Label}\\
Select the final topic label based on algorithm results.};

\draw[arrow] (s1) -- (s2);
\draw[arrow] (s2) -- (s3);
\draw[arrow] (s3) -- (s4);
\draw[arrow] (s4) -- (s5);
\end{tikzpicture}
\caption{Visual schema of the proposed methodology for TL, illustrating the main steps from input topic words to the selection of the final representative label.}
\label{fig:methodology}
\end{figure}
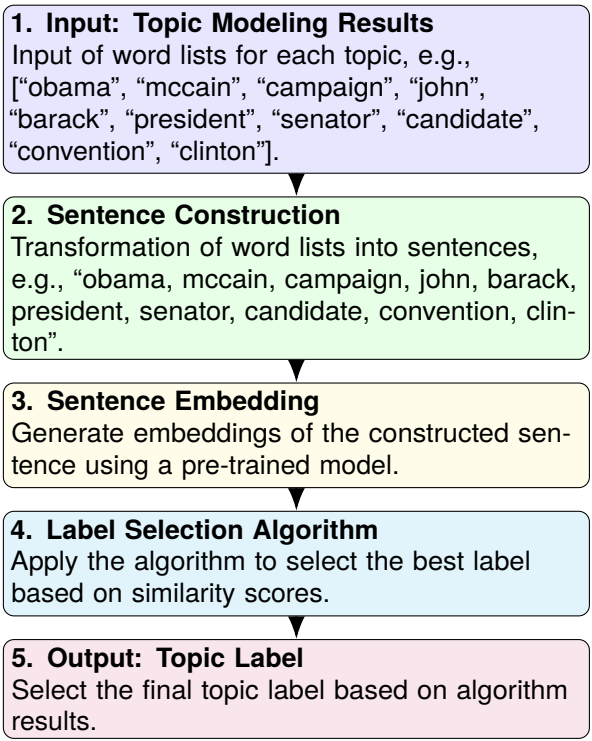

Our proposed methodology to address the problem of TL is divided into 5 parts, as shown in Figure~\ref{fig:methodology}. First, the input consists of a set of words representing the topic to be discovered, which corresponds to the results of TM. Using this input, we construct a sentence and transform it into its embedding (Section~\ref{sec:sentence-embedding}). Then, an algorithm specifically designed to extract the most suitable label for each topic set, is applied to generate the final representative label (Section~\ref{sec:labeling-algorithm}). Each component of the process is detailed in the following subsections.
\subsection{Sentence Construction and Embedding}
\label{sec:sentence-embedding}
We define the embedding of the sentences as the embedding obtained by encoding the words of a topic as a single sentence. Formally, let a topic be represented by a set of words
\[
T = \{ w_1, w_2, \dots, w_n \}.
\]
We construct a sentence by concatenating the words with commas and spaces:
\[
S = \text{}w_1, w_2, \dots, w_n\text{}
\]
and compute its embedding:
\[
E_{\text{topic}} = \textit{Embed}(S),
\]
where \textit{Embed} denotes a pre-trained embedding model. This embedding captures the semantic representation of the topic as a whole.

We adopt this method because the words representing a topic are semantically interdependent and must be considered together to deliver the intended meaning. For instance, as shown in Table~\ref{tab:dataset_examples}, in the second row, the word "president" alone does not clearly indicate that the topic concerns the President of the United States. However, when combined with related words such as "Barack", "Obama", "John", "McCain", "senator", and "Clinton", the topic becomes unambiguous. Similarly, the inclusion of the word "candidate" further reinforces the interpretation that the topic relates to U.S. presidential nominees. This observation motivates us to treat the set of topic words as a single sentence and compute its embedding accordingly, rather than embedding words separately.

To ensure robustness and identify the most suitable embedding model for our method, we evaluate a diverse set of pre-trained models, including those listed in Table~\ref{tab:embedding-model-params}. These models are selected based on three key criteria: they are open-source, have manageable parameter sizes (all under 400M), and demonstrate strong performance in benchmark evaluations, as reported by the \textit{Massive Text Embedding Benchmark}~\cite{huggingface_mteb}. This selection enables a comprehensive evaluation of embeddings that balance efficiency and accuracy, reinforcing the robustness of our method.
\begin{table}[!ht]
\begin{center}
\begin{tabularx}{\columnwidth}{|X|c|}
\hline
\textbf{Models} & \textbf{Parameters} \\
\hline
gtr-t5-large~\cite{ni2021large} & 335M \\
\hline
GIST-large-Embedding-v0~\cite{solatorio2024gistembed} & 335M \\
\hline
all-mpnet-base-v2\footnotemark[1] & 109M \\
\hline
GIST-small-Embedding-v0~\cite{solatorio2024gistembed} & 33M \\
\hline
all-MiniLM-L12-v2\footnotemark[3] & 33M \\
\hline
all-MiniLM-L6-v2\footnotemark[2] & 22M \\
\hline
GIST-all-MiniLM-L6-v2~\cite{solatorio2024gistembed} & 22M \\
\hline
\end{tabularx}
\caption{List of pre-trained embedding models considered in our evaluation, including their respective parameter sizes.}
\label{tab:embedding-model-params}
 \end{center}
\end{table}
\footnotetext[1]{\url{https://huggingface.co/sentence-transformers/all-mpnet-base-v2}}
\footnotetext[2]{\url{https://huggingface.co/sentence-transformers/all-MiniLM-L6-v2}}
\footnotetext[3]{\url{https://huggingface.co/sentence-transformers/all-MiniLM-L12-v2}}
Each pre-trained model produces its own topic embedding \(E_{\text{topic}}^{}\), which serves as a reference in the subsequent experiments. In particular, individual candidate word embeddings \(E_{c_{w_i}}^{} = \textit{Embed}^{}(c_{w_i})\) from each experiment are compared to the corresponding \(E_{\text{topic}}^{}\) to determine the most representative word for the topic. By evaluating multiple embedding models, we aim to compare their effectiveness and select the one that yields the most accurate and semantically meaningful labels in our setting.
\subsection{Label Selection Algorithm}
\label{sec:labeling-algorithm}
In this subsection, we present two algorithms that were adopted in our method. These algorithms are closely related; however, the first algorithm relies solely on the set of TM words, whereas the second incorporates graph-based information. We hypothesize that integrating the graph will add greater semantic richness to the set of words and help produce more representative and accurate topic labels. Nevertheless, there is also a possibility that the graph may introduce additional noise, making it unnecessary if the TM words alone are sufficient. Therefore, we designed two separate algorithms to evaluate both scenarios and later compare their performance to determine which achieves the best results.
\subsubsection{Algorithm~\ref{alg:Algorithm-A}: Direct Similarity Labeling (DSL)}
We compare each single candidate word embedding \(E_{c_{w_i}}^{}\) from the original word set that represents the TM results to \(E_{\text{topic}}\) and select the word with the highest similarity as the topic label. The objective of this method is twofold: (i) to evaluate how well individual embeddings and the aggregated topic embedding perform in identifying the most representative label, and (ii) to validate our general assumption that the most representative label is the word whose embedding is closest to the topic embedding \(E_{\text{topic}}\).
\begin{algorithm}[!ht]
\caption{Direct Similarity Labeling (DSL)}
\label{alg:Algorithm-A}
\KwIn{Sentence embedding $E_{\text{topic}}$, Topic Keywords $\{W_1, W_2, \dots, W_n\}$}
\KwOut{Final label $L^*$}

\ForEach{Topic keyword $W_i$}{
    Compute embedding $E_{w_i}$ for $W_i$\;
    Compute cosine similarity $S_i = \cos(E_{\text{topic}}, E_{w_i})$\;
}
Select topic label $L^* = W_j \;\; \text{where } j = \arg\max_{i} S_i$\;
\Return $L^*$
\end{algorithm}
\subsubsection{Algorithm~\ref{alg:Algorithm-B}: Graph-Enhanced Labeling (GEL)}
\paragraph{Graph Generation Process}
\label{sec:graph-generation}
For graph generation, we adopted the method proposed by Akhil~\emph{et. al}~\cite{chaudhary2024top2label}. Starting from the words output by TM, we use the ConceptNet API~\cite{liu2004conceptnet} to construct the most highly connected graph achievable. The intuition is that words belonging to the same topic are likely to be closely related within the ConceptNet knowledge graph. We chose ConceptNet~\cite{liu2004conceptnet} as our external knowledge source because it is freely available, contains rich semantic relations, and has shown strong results in previous studies~\cite{chaudhary2024top2label}. Our procedure involves two steps: (1) querying each word individually to retrieve its associated concepts and (2) iteratively expanding the graph by querying the neighbors of these concepts. We limited the expansion to three hops, following the findings of Akhil~\emph{et. al}~\cite{chaudhary2024top2label} that three iterations are sufficient to obtain a well-connected graph without introducing excessive noise.

\begin{figure*}[!ht]
\centering
\includegraphics[width=\textwidth]{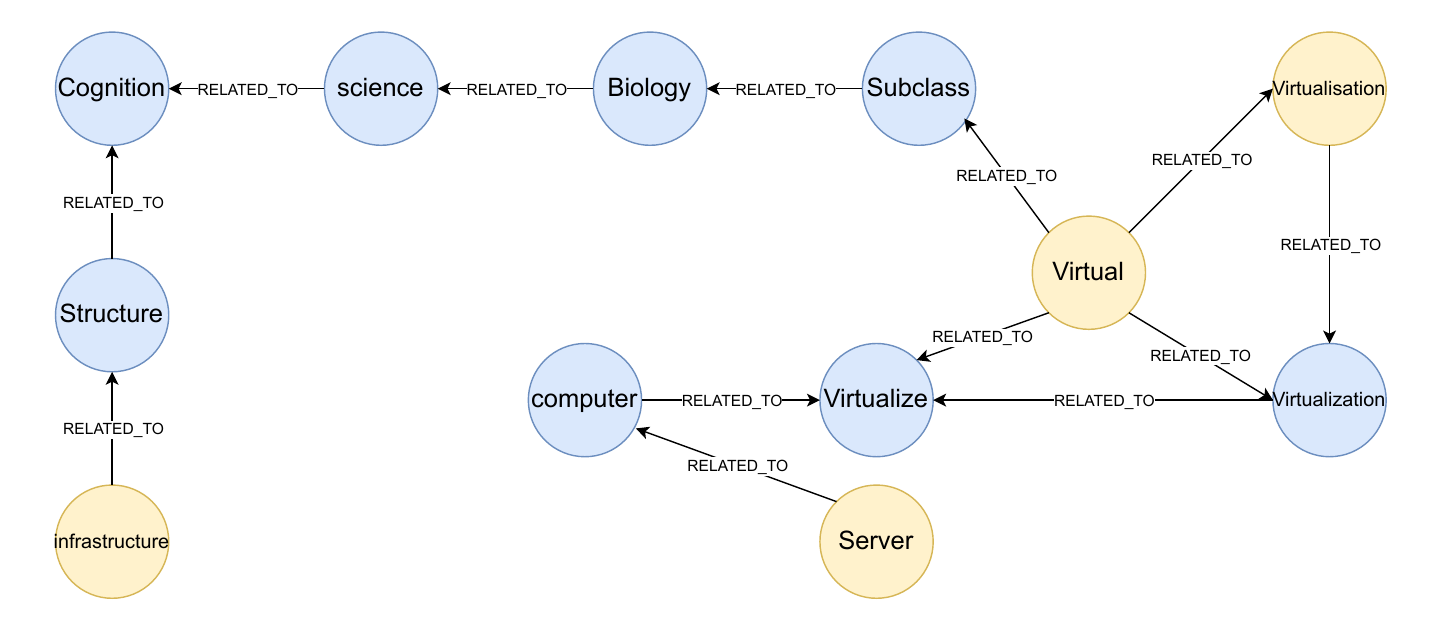} 
\caption{Visualization of how topic words (e.g., “server”, “infrastructure”, “virtualization”, and “virtual”) become interconnected within ConceptNet after iterative expansion, resulting in a well-connected graph.}
\label{fig:connection_graph}
\end{figure*}

The motivation behind this method is our hypothesis that the words generated by the TM represent a coherent theme. Therefore, when mapped into a concept graph, these words should eventually become interconnected after a few hops. This process allows us to construct a connected graph that captures semantic links among topic words. As illustrated in Figure~\ref{fig:connection_graph}, words from the first topic (e.g., “server”, “infrastructure”, "virtualization", and “virtual”) become connected through ConceptNet, enriching the vocabulary and providing a clearer understanding of the relationships between them.
\paragraph{Algorithm description}Formally, We compare each single candidate word embedding \(E_{c_{w_i}}^{}\) from \textit{the nodes of the constructed graph} to \(E_{\text{topic}}\), and select the word with the highest similarity as the topic label. The objective of this algorithm is to evaluate whether incorporating the graph structure improves label quality. The graph provides an enriched vocabulary that is closely related to the topic, allowing candidate labels to be drawn not only from the original set of words but also from graph-connected nodes.
\begin{algorithm}[!ht]
\caption{Graph-Enhanced Labeling (GEL)}
\label{alg:Algorithm-B}
\KwIn{Sentence embedding $E_{\text{topic}}$, Topic Keywords $\{W_1, W_2, \dots, W_n\}$}
\KwOut{Final topic label $L^*$}
Generate a ConceptNet-associated graph $G$ from the topic keywords$\{W_1, W_2, \dots, W_n\}$\;
Compute Embedding of all nodes of $G$ to obtain node embeddings $\{En_1, En_2, \dots, En_m\}$\;
\ForEach{node $n_j$ with embedding $En_j$ in $G$}{
    Compute cosine similarity $S_j = \cos(E_{\text{topic}}, En_j)$\;
}
Select Topic label $L^* = n_{k} \;\; \text{where } k = \arg\max_{j} S_j$
\Return $L^*$
\end{algorithm}
\section{Experiments, Results and Discussions}
\label{sec:Experiments}
In this section, we present and analyze the experimental results obtained from the proposed algorithms, DSL and GEL. The section is divided into two main parts. The first part evaluates DSL and GEL on the Topic\_Bhatia dataset, allowing comparison with prior studies that have shown strong performance in topic labeling for this corpus~\cite{chaudhary2024top2label,alokaili2020automatic}. The second part reports results on the 20 Newsgroups dataset, demonstrating the generalizability of our method across different domains and enabling comparison with previous works on this benchmark~\cite{ramon2023automatic}.

In addition to evaluating our algorithms, we apply a large pre-trained model specifically fine-tuned for TL, BART-TL~\cite{popa2021bart}, to the Topic\_Bhatia dataset. While benchmark results from prior studies are already reported, BART-TL has not previously been applied to this dataset. Therefore, we conduct this experiment to assess the performance of its two variations (BART-TL-ng and BART-TL-all) directly and to compare it with our algorithms as well as with the reported benchmarks. This setup allows us to evaluate both the effectiveness of our method and the potential advantages of applying a pre-trained model in this context~\cite{popa2021bart}.

All datasets, experimental setups, and results are available in a repository\footnote{\url{https://github.com/SalmaMekaoui/Enhancing_LLms_through_knowledge_graphs}}.
\subsection{Experiments on Topic\_Bhatia Dataset}
The evaluation of the quality of the generated labels is conducted using BERTScore~\cite{zhang2019bertscore}. BERTScore computes the similarity between predictions and references based on contextual embeddings, and has been shown to correlate strongly with human judgments, as demonstrated in its original paper~\cite{zhang2019bertscore}. One of the main limitations of traditional TL evaluation methods is their reliance on exact word matches, such as the accuracy, precision, recall, F1 score, and macro F1 score, which overlook semantically related labels~\cite{mekaoui2025systematic}. To address this, we adopt BERTScore since it accounts for semantic similarity rather than requiring exact lexical overlap. Furthermore, as the baselines also employed BERTScore for evaluation, using the same metric ensures a fair and consistent comparison across all methods.
\subsubsection{Dataset Description}
The Topic\_Bhatia dataset contains 228 topics, each associated with 19 candidate labels. The topics are drawn from four domains: blogs articles, English language books, New York Times news articles, and PubMed biomedical abstracts~\cite{lau2011automatic}. Human ratings for these labels were collected through a crowdsourcing task on Amazon Mechanical Turk (MTurk), where annotators evaluated labels on a scale from 0 (lowest) to 3 (highest). Only labels with an average rating of 2 or above were retained, resulting in a refined dataset of 219 topics and 1,156 topic label pairs, compared to the original 4,332 combinations. Examples of this dataset are mentioned in Table~\ref{tab:dataset_examples}.
\begin{table*}[ht]
\centering
\begin{tabular}{|p{2.5cm}|p{6cm}|p{6.25cm}|}
\hline
\textbf{Dataset} & \textbf{Example Terms} & \textbf{Possible Labels} \\
\hline \hline
\multirow{2}{*}{Topic\_Bhatia} 
 & vmware, server, virtual, oracle, update, virtualization, application, infrastructure, management, microsoft & cloud computing, microsoft exchange server, vmware, web application, virtualization, operating system \\ \cline{2-3}
 & obama, mccain, campaign, john, barack, president, senator, candidate, convention, clinton & conservative, democrat, democratic party (US), presidential nominee, bill clinton \\ \hline \hline
\multirow{2}{*}{20 Newsgroups} 
 & game, team, play, hockey, player, win, goal, season, fan, playoff & sport hockey \\ \cline{2-3}
 & god, religion, atheist, moral, claim, point, objective, good, belief, argument & religion atheism \\ \hline
\end{tabular}
\caption{Example topic terms and corresponding possible labels for the two datasets used in our study: \textit{Topic\_Bhatia} and \textit{20 Newsgroups}.}
\label{tab:dataset_examples}
\end{table*}

\subsubsection{Results and Discussion}
As shown in Table~\ref{tab:12_experiments_results}, the proposed algorithms achieve high performance across all BERTScore metrics. Prior to these experiments, the highest recorded F1 score on this dataset was 0.936, achieved by Top2Label~\cite{chaudhary2024top2label} (see Table~\ref{tab:123_experiments_results}). In comparison, DSL reached an F1 score of 0.955 using the \texttt{all-MiniLM-L6-v2} embedding, demonstrating that, in some cases, a single word from the topic modeling results is sufficient to identify the most accurate topic label. GEL also achieved a top F1 score of 0.955, this time using \texttt{gtr-t5-large}. Notably, GEL performed consistently well across almost all embeddings, indicating that incorporating graph-based relationships between words enriches the contextual information and produces labels that are semantically closer to the gold standard.

These results suggest that both DSL and GEL provide significant improvements over existing methods, while the main difference between DSL and GEL lies in the balance between precision and recall. In DSL, precision and recall are more closely aligned, suggesting that the candidate label captures most of the reference information without introducing irrelevant content. In GEL, precision exceeds recall, indicating that while the predicted label is semantically accurate, it covers only a limited portion of the reference. Although, the numerical difference is relatively small (0.02).

Turning to pretrained models, as presented in Table~\ref{tab:123_experiments_results}, BART-TL-ng achieved an F1 score of 0.892. Across all metrics, the results of this experiment did not surpass those of DSL and GEL, nor did they exceed the benchmark. This highlights limitations of pre-trained models in generating semantically representative labels for TM results.

Overall, these results indicate that using relatively simple methods based on the TM words alone is sufficient to achieve high-quality labels, and there is no need for more complex or computationally costly approaches. Although DSL and GEL achieved very similar scores, determining which method performs better would require a manual comparison between the predicted and gold labels, which remains resource-intensive. Furthermore, the Topic\_Bhatia dataset includes multiple gold reference labels for each topic word set as shown in Table~\ref{tab:dataset_examples}. Accordingly, the evaluation algorithm computes BERTScore between each generated label and all reference labels separately, selecting the highest similarity score as the final evaluation value. This setup helps explain the relatively high scores obtained by both DSL and GEL.

To further validate our method, we applied the same set of experiments to the 20 Newsgroups dataset. The results from this evaluation are reported in Section~\ref{sec:20_news}.
\begin{table*}[ht]
\centering
\begin{tabular}{|p{1.8cm}|p{3.5cm}|c|c|c|}
\hline
\textbf{Experiment} & \textbf{Embedding Method} & \textbf{BERT Precision} & \textbf{BERT Recall} & \textbf{BERT F1} \\ \hline
\multirow{7}{*}{DSL} 
 & all-MiniLM-L6-v2 & \textbf{0.958} & \textbf{0.953} & \textbf{0.955} \\
 & all-MiniLM-L12-v2 & 0.948 & 0.944 & 0.945 \\
 & all-mpnet-base-v2 & 0.943 & 0.944 & 0.943 \\
 & gtr-t5-large & 0.944 & 0.943 & 0.942 \\
 & GIST-small-Embedding-v0 & 0.951 & 0.951 & 0.951 \\
 & GIST-large-Embedding-v0 & 0.940 & 0.943 & 0.941 \\
 & GIST-all-MiniLM-L-v2 & 0.952 & 0.949 & 0.949 \\ \hline \hline
\multirow{7}{*}{GEL} 
 & all-MiniLM-L6-v2 & \textbf{0.964} & 0.945 & 0.954 \\
 & all-MiniLM-L12-v2 & 0.961 & 0.944 & 0.952 \\
 & all-mpnet-base-v2 & 0.961 & 0.945 & 0.952 \\
 & gtr-t5-large & 0.963 & \textbf{0.948} & \textbf{0.955} \\
 & GIST-small-Embedding-v0 & 0.955 & 0.939 & 0.946 \\
 & GIST-large-Embedding-v0 & 0.961 & 0.942 & 0.951 \\
 & GIST-all-MiniLM-L6-v2 & 0.956 & 0.937 & 0.946 \\ \hline
\end{tabular}
\caption{Performance comparison of the application of the adopted method. Experiments with DSL and GEL, corresponding to algorithms~\ref{alg:Algorithm-A} and~\ref{alg:Algorithm-B}, report results across multiple embedding models.}
\label{tab:12_experiments_results}
\end{table*}

\begin{table*}[ht]
\centering
\begin{tabular}{|p{3.2cm}|p{4.5cm}|c|c|c|}
\hline
\textbf{Category} & \textbf{Method} & \textbf{BERT Precision} & \textbf{BERT Recall} & \textbf{BERT F1} \\ \hline
\multirow{3}{*}{Baselines} 
 & Alokaili topics\_bhatia & 0.919 & 0.926 & 0.919 \\
 & Alokaili topics\_bhatia\_tfidf & 0.930 & 0.933 & 0.929 \\
 & Top2Label\_all\_relations & \textbf{0.935} & \textbf{0.937} & \textbf{0.936} \\ \hline
\multirow{2}{*}{BART-TL Benchmark} 
 & BART-TL-all & 0.873 & 0.900 & 0.886 \\ 
 & BART-TL-ng & \textbf{0.880} & \textbf{0.905} & \textbf{0.892} \\ \hline
\multirow{2}{*}{Proposed Method} 
 & DSL(all-MiniLM-L6-v2) & 0.958 & \textbf{0.953} & \textbf{0.955} \\ 
 & GEL(gtr-t5-large) & \textbf{0.963} & 0.948 & \textbf{0.955} \\ \hline
\end{tabular}
\caption{Performance comparison of baseline, proposed, and pretrained methods. For the proposed algorithms, DSL and GEL, only the results corresponding to the best-performing embeddings are reported. The table also includes a pretrained benchmark model (BART-TL) for comparison.}
\label{tab:123_experiments_results}
\end{table*}

\subsection{20 Newsgroups Dataset}
\label{sec:20_news}
For the 20 Newsgroups dataset, we evaluated the quality of the generated labels using the cosine similarity metric. This evaluation method differs from the previous experiments because our primary objective here is to directly compare our method with an established benchmark~\cite{ramon2023automatic}. Virginia~\emph{et al.}~\cite{ramon2023automatic} employed the same dataset and assessed multiple conversational models to determine their effectiveness in producing semantically coherent labels based on cosine similarity. The authors find out that the best method was chatGPT 3.5 with a cosine similarity score of 0.655~\cite{ramon2023automatic}. 

To ensure a fair and consistent comparison, we adopted the same evaluation strategy and applied the same embedding method, namely \texttt{all-MiniLM-L6-v2}, for computing cosine similarity scores.
Another objective of this experiment is to evaluate the impact of DSL and GEL across multiple datasets. The experiments on the Topic\_Bhatia dataset were unable to provide an unbiased and conclusive comparison of DSL and GEL. This limitation motivated us to extend the evaluation to additional datasets to better assess which algorithm produces the most effective results.
\subsubsection{Dataset Description}
This dataset consists of more than 20,000 posts organized into 20 distinct newsgroups and is widely used in document clustering research. In this work, we use a version where each of the 20 topics is represented by its label and a set of the 10 most representative words. Examples of this dataset are presented in Table~\ref{tab:dataset_examples}.
\subsubsection{Results and Discussion}
\label{Results_20_news}
\begin{figure*}[!ht]
\begin{center}
\includegraphics[width=0.8\textwidth]{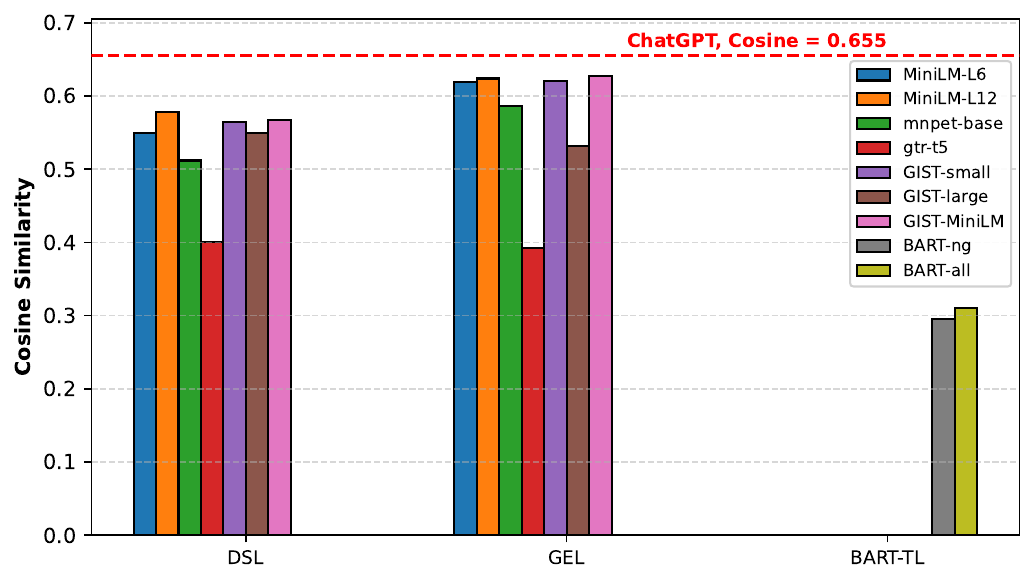}
\caption{Cosine similarity scores across the application adopted method on the 20 Newsgroups dataset. The red dashed line represents the benchmark (ChatGPT, 10 words).}
\label{fig:cosine_similarity_results}
\end{center}
\end{figure*}

From the results illustrated in Figure~\ref{fig:cosine_similarity_results}, it is evident that GEL outperforms both DSL and Bart-TL. This suggests that incorporating graphs provides significantly better results compared to using only the original set of words or a pre-trained model. Furthermore, although DSL and GEL do not surpass the benchmark performance achieved by ChatGPT (cosine similarity = 0.655), they still achieve strong results. Specifically, GEL achieves 0.627 with the GIST-all-MiniLM-L6-v2 embedding, while DSL achieves a cosine similarity of 0.578 using the all-MiniLM-L12-v2 embedding. Notably, these two experiments, which are highly interpretable and explainable, outperform BART-TL, which achieved a very low cosine similarity of 0.31(as shown in Table~\ref{tab:combined_cosine_exp123}), highlighting the effectiveness of simple and explainable methods. Moreover, these results are obtained with a relatively simple architecture and minimal computational resources. 

For the \textit{Topic\_Bhatia} dataset, each topic could be associated with multiple possible labels, which made it challenging to determine a consistent gold standard for evaluation. As for the 20 Newsgroups dataset, despite its well-established reputation in classification tasks, it remains quite limited for topic labeling purposes. The number of topic modeling keywords and labels is restricted to only 20, which constrains the diversity and scalability of experiments.
\begin{table}[ht]
\centering
\small
\caption{Cosine similarity scores across the application adopted method on the 20 Newsgroups dataset.}
\label{tab:combined_cosine_exp123}
\begin{tabular}{|p{2.5cm}|c|c|c|c|c|}
\hline
\textbf{Embedding} & \rotatebox{90}{\textbf{DSL}} & \rotatebox{90}{\textbf{GEL}} & \rotatebox{90}{\textbf{BART-TL-all}} & \rotatebox{90}{\textbf{BART-TL-ng}} \\ \hline
\textbf{No Embedding} & - & - & 0.310 & 0.295 \\ \hline
\textbf{all-MiniLM-L6-v2} & 0.550 & 0.619 & - & - \\ \hline
\textbf{all-MiniLM-L12-v2} & \textbf{0.578} & 0.624 & - & - \\ \hline
\textbf{all-mpnet-base-v2} & 0.512 & 0.587 & - & - \\ \hline
\textbf{gtr-t5-large} & 0.401 & 0.392 & - & - \\ \hline
\textbf{GIST-small-Embedding-v0} & 0.564 & 0.620 & - & - \\ \hline
\textbf{GIST-large-Embedding-v0} & 0.550 & 0.532 & - & - \\ \hline
\textbf{GIST-all-MiniLM-L6-v2} & 0.567 & \textbf{0.627} & - & - \\ \hline
\end{tabular}
\end{table}

\section{Conclusion}
\label{sec:conclusion}
In this paper, we proposed two methods for topic labeling that assign representative labels to topic modeling outputs using only the topic modeling results themselves and their corresponding representative graphs. Experiments conducted on two datasets demonstrated strong performance in terms of BERTScore and cosine similarity when compared to both a fine-tuned pre-trained model for topic labeling and existing benchmarks.

In terms of F1 score, DSL and GEL achieved 0.955 using the \textit{all-MiniLM-L6-v2} and \textit{gtr-T5-large} embeddings, respectively, surpassing the highest previously reported benchmark of 0.936. GEL, which incorporates graph representations to enrich the topic word sets with semantically informative context, produced excellent results on the 20 Newsgroups dataset. Using the \textit{GIST-all-MiniLM-L6-v2} embedding model, it achieved a cosine similarity score that was only 0.028 lower than that of GPT-3.5, despite requiring significantly fewer computational resources.

These findings suggest that simple and interpretable methods, such as combining topic modeling results with graph-based representations, can outperform more complex pre-trained models in generating semantically meaningful topic labels, while remaining computationally efficient. Furthermore, results on the 20 Newsgroups dataset confirmed the robustness and generalizability of our approach across different domains.

Although the experiments yielded strong results, several limitations should be acknowledged regarding the size and nature of the datasets used. Future work could explore the creation of larger, high-quality labeled datasets specifically designed for the topic labeling task, especially considering the growing importance of topic labeling in extracting meaningful topics from large-scale text corpora.

Moreover, the promising results obtained through graph-based expansion of topic keywords open new research directions. Future studies could explore alternative knowledge graphs beyond ConceptNet, leverage graph structures to identify the most informative nodes, or apply advanced graph representation techniques, such as graph2vec, to derive more representative and semantically rich topic labels.
\section{Acknowledgements}
This publication has emanated from research conducted with the financial support of Taighde Éireann - Research Ireland under Grant number 18/CRT/6223.
\section{References}
\bibliographystyle{lrec2026-natbib}
\bibliography{lrec2026-example}
\bibliographystylelanguageresource{lrec2026-natbib}
\bibliographylanguageresource{languageresource}
\end{document}